\pdfoutput=1

\documentclass[11pt]{article}

\usepackage{ACL2023}

\usepackage{graphicx}

\usepackage{times}
\usepackage{latexsym}
\usepackage{amsmath}
\usepackage{amsthm}
\usepackage{hyperref}
\DeclareMathOperator*{\argmaxB}{argmax}

\usepackage[english]{babel}

\usepackage[T1]{fontenc}

\usepackage[utf8]{inputenc}
\DeclareUnicodeCharacter{57FA}{\`-}

\usepackage{microtype}

\usepackage{inconsolata}

\title{Confidence-Calibrated Ensemble Dense Phrase Retrieval}

\author{
  William Yang\\
  \texttt{wyyang@mit.edu} \\\And
  Noah Bergam \\
  \texttt{njb2154@columbia.edu} \\\And
  Arnav Jain \\
  \texttt{arnavj@ucla.edu }\\\And 
  Nima Sheikhoslami \\
  \texttt{nimasheikh96@gmail.com}
}
\begin{document}
\maketitle
\begin{abstract}
In this paper, we consider the extent to which the transformer-based Dense Passage Retrieval (DPR) algorithm, developed by \cite{karpukhin2020dense}, can be optimized without further pre-training. Our method involves two particular insights: we apply the DPR context encoder at various phrase lengths (e.g. one-sentence versus five-sentence segments), and we take a confidence-calibrated ensemble prediction over all of these different segmentations. This somewhat exhaustive approach achieves start-of-the-art results on benchmark datasets such as Google NQ and SQuAD. We also apply our method to domain-specific datasets, and the results suggest how different granularities are optimal for different domains.
\end{abstract}

\section{Introduction}

The passage retrieval problem, which is of central importance in search engine optimization and text analytics, entails the following: given a set of documents and a query, determine which document best answers the question. This document retrieval step is an important intermediary in the more involved problem of open-domain (extractive) question answering (OPQA), in which one is tasked with finding the specific substring in the relevant document that contains the answer to the question. In this paper, we focus on one of the state-of-the-art models for automated document retrieval: the transformer-based dense passage retrieval algorithm, developed by \cite{karpukhin2020dense}.

Our analysis is primarily motivated by the observation in \cite{lee2021phrase} that DPR can yield improved performance when applied to smaller chunks of text (e.g. sentences rather than paragraphs). This slight adaptation of DPR is known as \textit{dense phrase retrieval}. In our work, we view phrase length as a parameter which can be altered to create slightly different DPR models. One can view the embeddings from each phrase length as the ``expert" of a certain granularity, and our prediction proceeds in an ensemble fashion, where we choose the expert with the highest confidence. This motivates the use of confidence calibration, which we accomplish using temperature scaling of the softmax output \cite{guo2017calibration}.

We apply this method to a range of datasets, including both large canonical datasets (Google NQ, SQuAD) and smaller domain-specialized datasets relating to information retrieval in law, medicine, and scientific research, respectively.

\section{Background}

In the literature of IR, Okapi BM-25 often sets the baseline. Developed in the 1980s, this statistical model ranks the best-matching passages by scoring them according to a modified term-frequency inverse document-frequency (TF-IDF) framework. The principal limitation to this approach is its dependence on explicit term matches between the query and the context. In many cases, the correct context-query pair may have no words in common. For instance, they may use synonyms or inflections to strike at the same ideas. Such cases severely undermine the accuracy of BM-25.

A modern perspective on this problem, informed by the rise of neural models, may immediately look to dense word embeddings as a workaround. A good embedding, the logic goes, is supposed to capture semantics in a way that can pick up on analogies, inflections, and synonyms. However, there has been a lot of trouble in making these dense embedding frameworks outperform the classical bag-of-words methods in information retrieval. It was not until very recently that dense methods were found to consistently outperform the simpler BM-25 approach \cite{karpukhin2020dense}.

An important predecessor to DPR was \cite{lee2021phrase}’s Open Retrieval Question Answering system (ORQA). This algorithm aims for end-to-end IR and QA, by training a retriever and reader jointly. Importantly, they try to make the retriever work better by using an inverse-cloze task (ICT): given a sentence, predict its correct context. \cite{karpukhin2020dense} is typically credited with developing the first, canonically superior dense retrieval method. The method is remarkably simple: we train two separate encoders (no ICT pretraining!) for context and query. The loss function is simply the negative log likelihood of the positive passage. Similarity scores between question and passage embeddings is calculated via an inner product or a cosine similarity function.

\begin{table*}[h!]
\begin{center}
\begin{tabular}{cccc}
  &     \(\#\)questions-context pairs & avg. question length & avg. context length \\ \hline
 SQUAD & 11.9k & 10.0 & 128.5   \\
 NQ\(_\text{small}\) & 8.2k & 49.3  & 36479.6 \\
 PubmedQA & 61.2k  & 13.3 & 196.4  \\
 SCOTUS & 271  & 201.7 & 29409.9 \\
NFCorpus & 102  & 33.7 & 1598.1 \\
\hline
\end{tabular}
\caption{ Basic dataset details. Question and context length are measured in tokens.}
\end{center}
\label{fig:table1}
\end{table*}

\begin{figure*}[h]
    \centering
    \includegraphics[width=5cm]{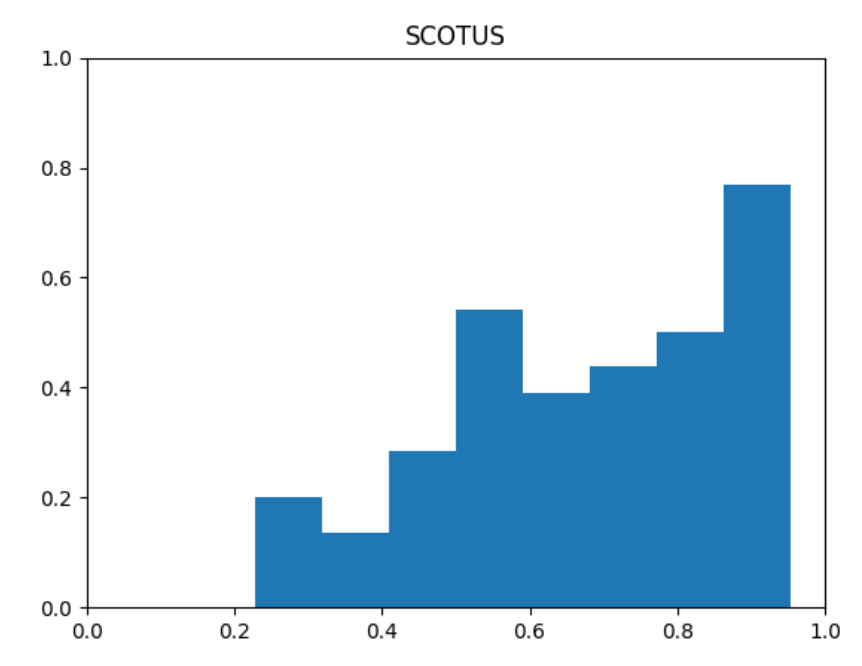}
    \includegraphics[width=5cm]{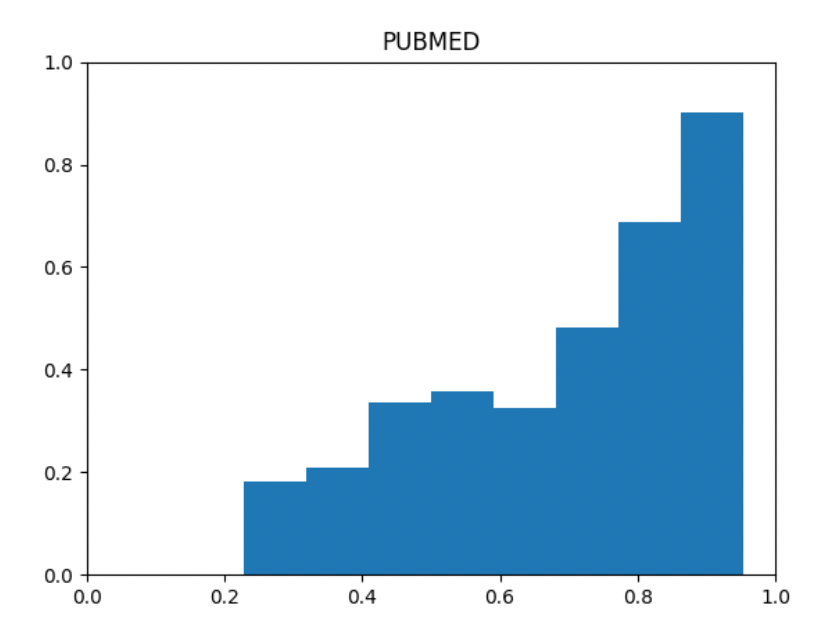}
    \includegraphics[width=5cm]{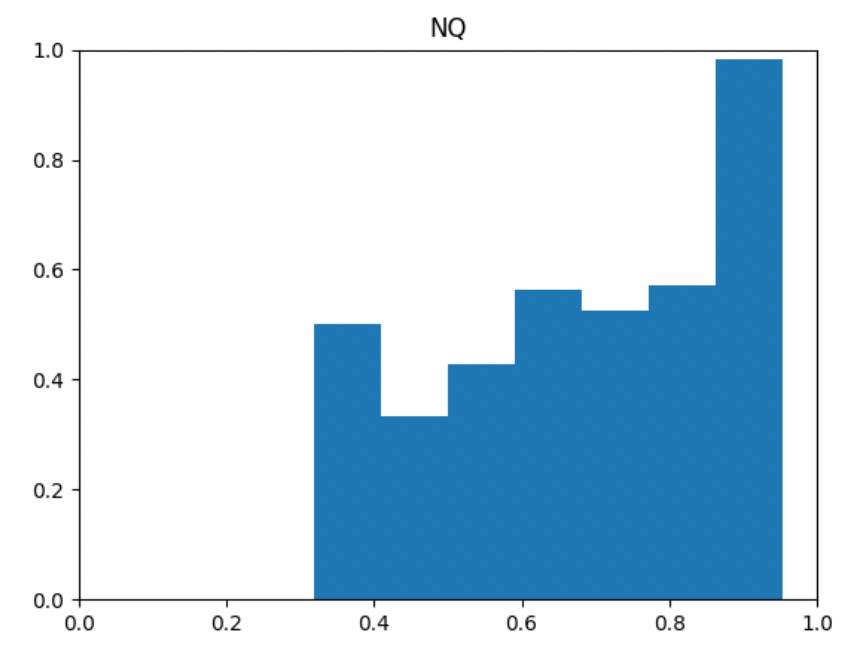}
    \caption{Expected calibration error (ECE) graphs of the original DPR model on SCOTUS, PubmedQA, and NFCorpus. The \(x\)-axis enumerates levels of confidence and the y-axis measures the accuracy of predictions made with that confidence.}
    \label{fig:my_label}
\end{figure*}

\cite{lee2021phrase} found that the dense passage retrieval could be improved in some cases if the passage of interest is split into “phrases” of a smaller length, and the question is considered with respect to each phrase. Others  have explored an ensemble approach to improve the DPR model. Rather than follow the methods of the original paper and train a DPR jointly on a number of different datasets like Google NQ, SQuAD, and TriviaQA, (Lee 2020) shows that model fusion can beat joint training when separate DPRs are trained on different datasets, and the answer of the most confident one is treated as the answer.

In \cite{wang-2002-ji}, single sentence embedding representations were used to improve performance of contrastive learning. In this method, instead of representing the entire passage as one unified embedding each sentence has its own allowed for more fine grained representations, which was shown to improve top 20 accuracy in major datasets such as NQ and TriviaQA.

\section{Methods}
\subsection{Dense Phrase Retrieval}
DPR's approach of encoding the entirety of a context passage can reduce the nuance that certain phrases provide to a limited size context encoding. One simple way to resolve this is to improve the granularity of contexts by splitting a context passage into smaller phrases and encoding them individually. \cite{lee2021phrase} pioneered this technique and called it dense phrase retrieval. Such a strategy has a number of additional benefits such as narrowing the search for specific answer phrases in extractive question answer tasks.

Extending upon the work of \cite{lee2021phrase}, we investigate whether dense phrase retrieval can be extended into an ensemble method, where different phrase lengths offer different predictions on which document has the best answer. Instead of segmenting phrases with fixed word lengths, we generate phrases from a context passage based off of the number of sentences, $N$. This method ensures that no important phrases that may contain a question answer may be split into two separate phrases avoiding any compromise of meaning.

Let us formalize the process of dense phrase retrieval. Typically, DPR predicts an answer passage $p_{\text{pred}}$ according to a maximum inner product search, which is defined as follows:
\[p_{\text{pred}}= \text{arg}\max_p \frac{E_p(p)^\top E_q(q)}{|E_p(p)|}\]
where $p$ is a passage, $q$ is a question, $E_p(p), E_q(q)$ are the passage encoding and question encoding respectively, and $E_p$ and $E_q$ are the passage and question encoders respectively.

For dense phrase retrieval, let $S(p)$ define the set of phrase encodings for passage $p$, $r$ be a given phrase in $S(p)$, and $E_r$ be a phrase encoder. The way we generate $p_{\text{pred}}$ is by first finding the maximum inner product score of a phrase in specific passage $p$ and then finding the passage with the highest score. In other words:
\[ p_{\text{pred}}=\text{arg} \max_{r \in S(p)} \frac{E_r(r)^\top E_q(q)}{|E_r(r)|}\]
Note the above is equivalent to finding the phrase with the highest inner product score with $q$.

In our research, we encoded passages according to three different phrase lengths: 1 sentence, 3 sentences, and 5 sentences. Due to limited computational resources, we did not train a custom phrase encoder and instead used a DPR encoder published by Facebook Research AI on Huggingface. 

\subsection{Model Uncertainty Fusion}
Let $M_n$ denote a dense phrase retriever model encoding phrases of sentence length $n$, and let $M_0$ denote the standard DPR model. Using multi-model fusion of the top 3 performing models among the set $S_M = \{M_1, M_3, M_5, M_0\}$ we can improve accuracy relative to DPR alone. 

In general, when we get answers from each model $M_n$, we want to select the answer from the model that is most confident of its prediction. Let $A_k$ denote the set of the $k$ highest scores from all the inner products produced from $E_q(q) \cdot E_r(r)$ from model $M_n$. Let $p_{\text{max}}$ be the score of the highest scoring phrase or passage of the model. In addition, let $T$ be some constant used for temperature scaling our softmax function.

We calculate confidence according to a softmax normalization.

\[\textbf{Conf}(M_n)=\frac{\exp({p_{\text{max}}/T})}{\sum\limits_{p \in A_k} \exp(p/T)}\]

Thus the best model $M^\ast$ and associated predicted passage $p_{\text{pred}}$ given query $q$ are as follows: $$M^\ast=\argmaxB_{M_i \in S_M} \textbf{Conf}(M_i)$$
$$p_{\text{pred}}=M^\ast(q) $$
We make use of temperature scaling with constant $T$ for fine tuning our softmax function. In order to calibrate our model properly, we use an ECE loss function \cite{guo2017calibration}. Here, we define $N$ different bins where bin $B_i$ consists of predictions with a confidence vales in range $\left[\frac{i-1}{N}, \frac{i}{N}\right)$.

Here we try to minimize the difference between the average confidence and accuracy of the bin. Instead of using absolute value difference as Guo et al. 2017 do, we take the mean squared difference as follows to maintain a differentiable function, 

$$\textbf{ECE}=\sum\limits_{i=1}^N\frac{|B_i|}{K} (\textbf{Conf}(B_i)-\textbf{Acc}(B_i))^2$$
where $K$ is the total number of entries.

The gradient of the \textbf{ECE} loss function relative to $T$ is as follows:
$$2\sum\limits_{i=1}^N\left(\frac{|B_i|}{K} (\textbf{Conf}(B_i)-\textbf{Acc}(B_i))\cdot \frac{1}{|B_i|}\sum\limits_{j=1}^{|B_i|}g_j\right)$$
where $g_j=\nabla_T \textbf{Conf}(M^\ast(p_{\text{pred}}))$, the gradient of the $j$th confidence score entry in bin $B_i$. The results of using this regression technique are reported in section 4.2.

\section{Results}

\subsection{Technical Details} 
We ran our experiments on eight Tesla K-80 GPUs, made available through an Amazon AWS server. We pre-processed and experimented on five datasets: the Stanford Question Answering Dataset or SQuAD \cite{rajpurkar-etal-2018-know}, Google Natural Questions or NQ \cite{kwiatkowski2019natural}, PubmedQA \cite{jin2019pubmedqa}, Supreme Court stance or SC-stance \cite{bergam2022legal}, and Nutrition Facts Corpus or NFCorpus \cite{boteva2016}. The first two are "generalist" in the sense that they answer a wide range of topics using the language of Wikipedia and webpages, while the latter three are "specialist" in the sense that they feature the language of a very specific expert group of people (i.e. medical and legal professions). Specifically, PubMed QA contains question context pairs sourced mainly from biological literature while NF-Corpus and SCOTUS contains data sourced from legal documents.

Since selecting the best passage from an entire dataset requires an amount of computation that scales with the size of that dataset, we instead choose to select passages from a representative batch of 30 passages, where one of them is correct, 10 are hard-negatives (as decided by the top 10 scoring passages from BM-25 excluding the correct passage), and 20 are random passages not already included.

For NQ, due to limited computational resources we trained on approximately $\frac{1}{40}$ of the train set and $\frac{1}{3}$ of the dev set. For evaluation below in table 5 we take the results from the training set. For general evaluation in section 4.3 we used results from the $\text{NQ}_{\text{small}}$ training set.

Before describing our results, it is useful to have a sense of the different scales and distributions presented by the datasets at hand. The sizes of each dataset along with the number of question-context pairs, and average question and context token length are reported in Table 1.

In this paper, due to our treatment of model fusion, we are also sensitive to the ideas of confidence calibration. As such, we also examine how the publicly available Huggingface DPR model is calibrated with respect to each of these datasets. We hypothesize that the more generalist datasets will be more calibrated than the specialist ones. 

\subsection{MUF for Phrases}
Table 2 reports the performance of the dense phrase retrieval models at various phrase lengths, as well as the model uncertainty fusion algorithm (MUF). Note that the top 3 models used for each instance of MUF varies depend on individual model performance on a given dataset.

\begin{table*}[h!]
\begin{center}
\begin{tabular}{|c|cccc|c|cc|}
\hline
 sentences & SQUAD & PubmedQA & SCOTUS & NFCorpus& tokens & NQ (train) & NQ (dev)\\ \hline
 1 & 53.4 & 77.8 & 66.2 & 30.8
 &75& 70.2 &  77.4   \\
 3 & 44.7 & 76.5 & 65.1 & 24.5
 &125& 69.0 & 74.0  \\
 5 & 47.0 & 72.7 & 61.4 & 21.9
 &250& 65.3 & 70.1 \\
MUF & 54.5 & 84.8 & 70.0 & 27.8
&MUF& 83.0 & 86.7 \\\hline
\end{tabular}
\caption{ F-1 scores of dense phrase retrieval evaluated on major datasets based on top 30 evaluation. }
\end{center}
\end{table*}



An important factor in MUF accuracy is deciding what initial starting temperature $T_0$ we should begin our regression with. Since gradient values tend to be small, we want relatively large step sizes. In our evaluation we used step sizes $\tau=10^2$. Below we present plots of accuracy for some datasets with varying values of $T_0$.

\begin{figure}[h!]
    \centering
    \includegraphics[width=6.5cm]{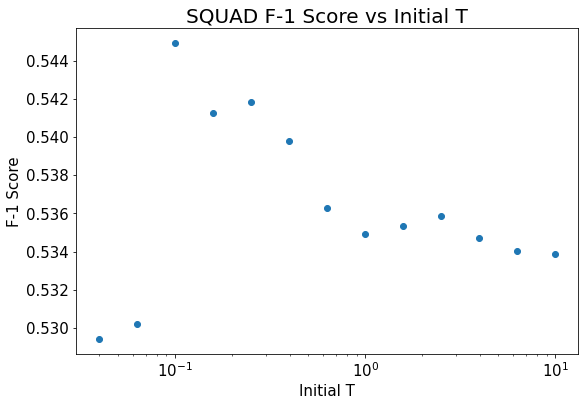}
    \includegraphics[width=6.5cm]{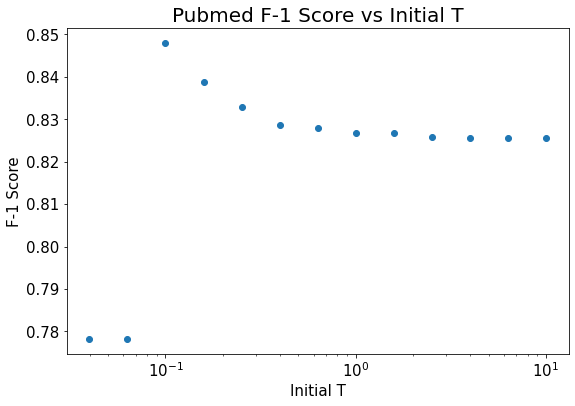}
    \caption{Varying accuracy depending on initial $T_0$ temperature parameter in the confidence calibration for SQuAD (top) and PubMed (bottom).}
    \label{fig:my_label}
\end{figure}
As per above, setting $T_0=0.1$ appears to yield the best accuracy. One possible explanation for why accuracy drops so significantly when $T_0<0.1$ is due to potential overflow errors when performing the softmax function, though further investigation needs to be done to properly analyze this.
\subsection{General Evaluation}
In our final round of experiments, we looked to compare MUF to established models such as DPR and BM-25 across multiple datasets spanning different subject domains. Our results are reported below in table 4. 

\begin{table*}[h!]
\centering
\begin{tabular}{c|ccccc}
  & NQ  & SQuAD & PubmedQA & SCOTUS & NFCorpus \\ \hline 
 BM25 & 30.3 & 75.4 & 85.6 & 86.9  & 25.3 \\
 DPR & 86.4 & 48.3 & 60.2 & 41.3 & 27.5 \\
 MUF & 86.7 & 54.4 & 84.0 & 70.0 & 27.8\\
\end{tabular}
\caption{ F-1 scores of DPR evaluated on major datasets, for all models except BM25 based on top 30 evaluation, where 9 are hard negatives and 20 are other randomly selected passages.  }
\end{table*}

\section{Conclusion}

The motivating question of this project––which found an affirmative answer in our experiments––is as follows: \textit{to what extent can we improve the publicly available DPR algorithm, without re-training or pre-training the entire model?} This is an important question because the default implementation is often the go-to for NLP practitioners. It is important to test that default's limits and see what kind of insights or improvements we can make with relatively simple, explainable adjustments. 

This paper sheds light on the importance of understanding the contexts of datasets and choosing model optimization strategies accordingly. For instance, when adapting a general DPR model to a more specialized domain, we can gain a good sense of how much adaptation it needs from looking at the calibration graph. 

We envision many potential improvements to CCE. Firstly, training a custom phrase encoder should yield significant improvements in the performance of individual dense phrase retriever models. Generating phrase datasets to use in batch training during meta learning could potentially increase performance in specific domains. Future work should consider making use of query side fine-tuning as proposed in \cite{lee2021phrase} to improve results.

Furthermore, it should be noted that the computational overhead of CCE is flexible. For example, in resource-limited settings where a high accuracy is crucial, the phrase length of each sub-model can be increased, which will lead to only a small magnitude of increase in computation time needed relative to that of the baseline while giving better accuracies.  On the other hand, more resource-rich settings can employ more fine grained phrase lengths in their submodels allowing for greater accuracies.


\section*{Acknowledgements}
This work was supported by the NLMatics 2022 summer internship program. The authors would like to thank Ambika Sukla for organizing this opportunity and providing invaluable guidance throughout the research process. 
\bibliographystyle{acl_natbib}
\bibliography{acl2023}

\end{document}